\documentclass[11pt,a4paper]{article}
\usepackage[hyperref]{naaclhlt2019}
\usepackage{times}
\usepackage{latexsym}

\usepackage{url}
\usepackage{booktabs}       
\usepackage{amsfonts}       
\usepackage{nicefrac}       
\usepackage{microtype}      
\usepackage{amsmath}
\usepackage{bbm}
\usepackage{color}
\usepackage{multirow}
\usepackage{graphicx}
\usepackage{caption}
\usepackage{subcaption}
\usepackage{amssymb}
\usepackage{enumitem}
\usepackage{hhline}
\usepackage[symbol]{footmisc}

\renewcommand{\eqref}[1]{Eq.~(\ref{#1})}

\renewcommand{\cite}{\citep}

\usepackage{pifont}
\newcommand{\cmark}{\ding{51}}%

\iffalse
\newcommand\eugeneie[1]{}
\newcommand\jason[1]{}
\newcommand\vihan[1]{}
\newcommand\rsents[1]{}
\newcommand\harsh[1]{}
\else
\newcommand\eugeneie[1]{[\textcolor{blue}{EI: {#1}}]}
\newcommand\jason[1]{[\textcolor{red}{JB: {#1}}]}
\newcommand\vihan[1]{[\textcolor{blue}{VJ: {#1}}]}
\newcommand\rsents[1]{[\textcolor{teal}{HH: {#1}}]}
\newcommand\harsh[1]{[\textcolor{orange}{HM: {#1}}]}
\fi

\newcommand\eg{e.g.}
\newcommand{\friedaug}{\texttt{Fried-Augmented}}

\aclfinalcopy 

\title{Multi-modal Discriminative Model for Vision-and-Language Navigation}

\author{Haoshuo Huang\thanks{\ Authors contributed equally.} \quad 
        Vihan Jain\footnotemark[\value{footnote}] \quad 
        Harsh Mehta \quad 
        Jason Baldridge \quad 
        Eugene Ie \\
  Google AI Language \\
  {\tt \{haoshuo, vihan, harshm, jridge, eugeneie\}@google.com}
  }

\begin{document}

\maketitle

\begin{abstract}
    
Vision-and-Language Navigation (VLN) is a natural language grounding task where agents have to interpret natural language instructions in the context of visual scenes in a dynamic environment to achieve prescribed navigation goals.
Successful agents must have the ability to parse natural language of varying linguistic styles, ground them in potentially unfamiliar scenes, plan and react with ambiguous environmental feedback.
Generalization ability is limited by the amount of human annotated data. In particular, \emph{paired} vision-language sequence data is expensive to collect.
We develop a discriminator that evaluates how well an instruction explains a given path in VLN task using multi-modal alignment.
Our study reveals that only a small fraction of the high-quality augmented data from \citet{Fried:2018:Speaker}, as scored by our discriminator, is useful for training VLN agents with similar performance on previously unseen environments.
We also show that a VLN agent warm-started with pre-trained components from the discriminator outperforms the benchmark success rates of 35.5 by 10\% relative measure on previously unseen environments.

\end{abstract}


\section{Introduction}
\label{sec:introduction}

There is an increased research interest in the problems containing multiple modalities \cite{Yu:2013:VideoCaptioning,Chen:2015:MSCOCO,Vinyals:2017:CaptioningMSCOCO,Harwath:2018:VisualAudio}. The models trained on such problems learn similar representations for related concepts in different modalities. Model components can be pretrained on datasets with individual modalities, the final system must be trained (or fine-tuned) on task-specific datasets \cite{Girshick:2014:Finetuning,Zeiler:2014:Pretraining}.

In this paper, we focus on vision-and-language navigation (VLN), which involves understanding visual-spatial relations as described in instructions written in natural language. In the past, VLN datasets were built on virtual environments, with \citet{Macmahon06walkthe} being perhaps the most prominent example. More recently, challenging photo-realistic datasets containing instructions for paths in real-world environments have been released \cite{Anderson:2018:VLN,DBLP:journals/corr/abs-1807-03367,ChenTouchdown2018}. Such datasets require annotations by people who follow and describe paths in the environment. Because the task is quite involved--especially when the paths are longer--obtaining human labeled examples at scale is challenging. For instance, the Touchdown dataset \cite{ChenTouchdown2018} has only 9,326 examples of the complete task. Others, such as \citet{Cirik:2018:StreetView} and \citet{streetlang} side-step this problem by using formulaic instructions provided by mapping applications. This makes it easy to get instructions at scale. However, since these are not \emph{natural} language instructions, they lack the quasi-regularity, diversity, richness and errors inherent in how people give directions. More importantly, they lack the more interesting connections between language and the visual scenes encountered on a path, such as \textit{head over the train tracks, hang a right just past a cluster of palm trees and stop by the red brick town home with a flag over its door}.

In general, the performance of trained neural models is highly dependent on the amount of available training data. Since human-annotated data is expensive to collect, it is imperative to maximally exploit existing resources to train models that can be used to improve the navigation agents. For instance, to extend the Room-to-Room (R2R) dataset \cite{Anderson:2018:VLN}, \citet{Fried:2018:Speaker} created an augmented set of instructions for randomly generated paths in the same underlying environment. These instructions were generated by a speaker model that was trained on the available human-annotated instructions in R2R. Using this augmented data improved the navigation models in the original paper as well as later models such as \citet{Wang:2018:RCM}. However, our own inspection of the generated instructions revealed that many have little connection between the instructions and the path they were meant to describe, raising questions about what models can and should learn from noisy, automatically generated instructions.

We instead pursue another, high precision strategy for augmenting the data. Having access to an environment provides opportunities for creating instruction-path pairs for modeling alignments. In particular, given a path and a navigation instruction created by a person, it is easy to create incorrect paths by creating permutations of the original path. For example, we can hold the instructions fixed, but reverse or shuffle the sequence of perceptual inputs, or sample random paths, including those that share the start or end points of the original one. Crucially, given the diversity and relative uniqueness of the properties of different rooms and the trajectories of different paths, it is highly unlikely that the original instruction will correspond well to the mined negative paths.

This negative path mining strategy stands in stark contrast with approaches that create new instructions. Though they cannot be used to directly train navigation agents, negative paths can instead be used to train discriminative models that can assess the fit of an instruction and a path. As such, they can be used to judge the quality of machine-generated extensions to VLN datasets and possibly reject bad instruction-path pairs. More importantly, the components of discriminative models can be used for initializing navigation models themselves and thus allow them to make more effective use of the limited positive paths available.

\begin{figure}
    \centering
     \includegraphics[clip, width=\linewidth]{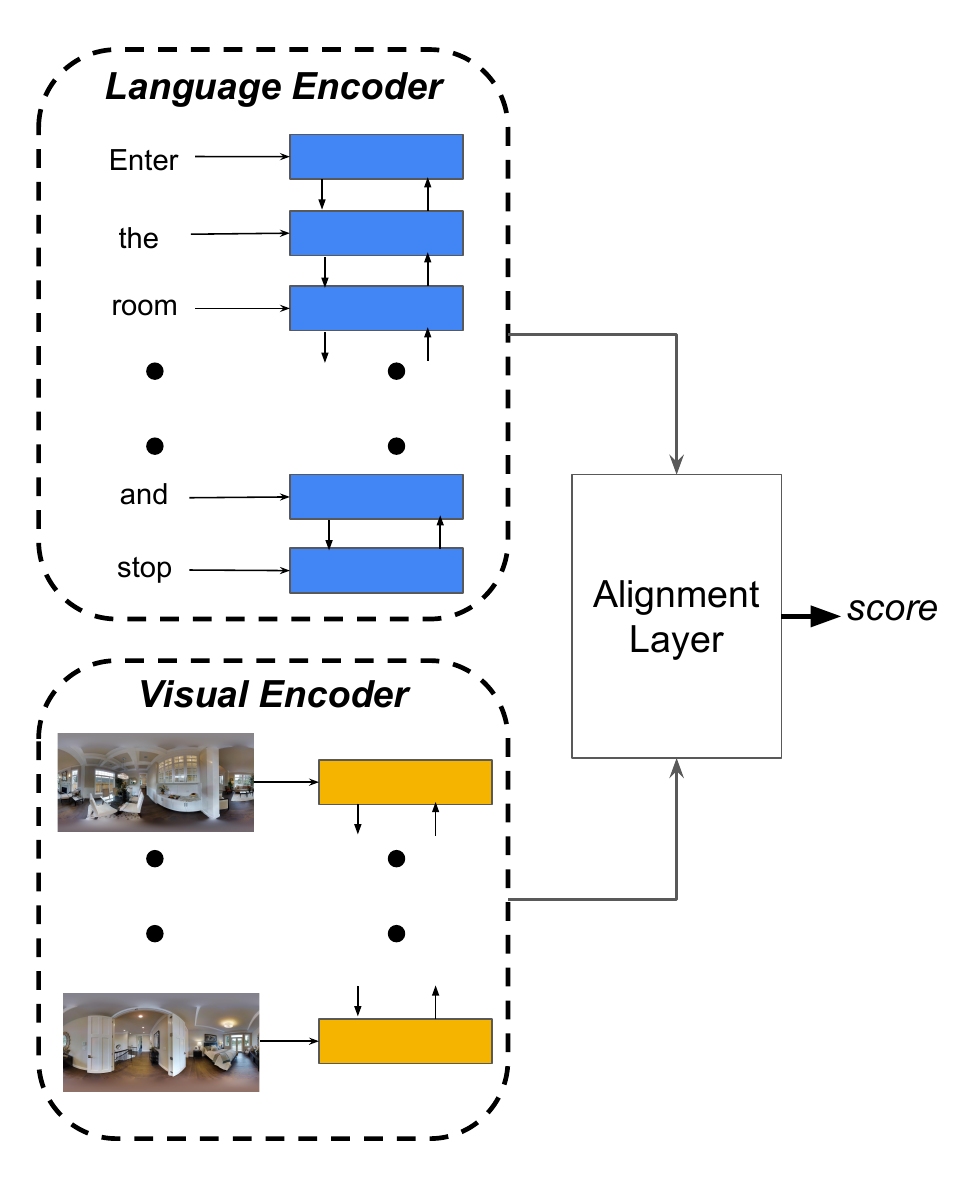}
    \qquad
    \scriptsize
    \caption{Overview of the discriminator model structure. Alignment layer corresponds to Eq.\ref{eq:A_compute},\ref{eq:A_softmax},\ref{eq:A_softmin}}

    \label{fig:deoverview}
\end{figure}

We present four main contributions. First, we propose a \textit{discriminator} model (Figure \ref{fig:deoverview}) that can predict how well a given instruction explains the paired path. We list several cheap negative sampling techniques to make the discriminator more robust. Second, we show that only a small portion of the augmented data in \citet{Fried:2018:Speaker} are high fidelity. Including just a small fraction of them in training is sufficient for reaping most of the gains afforded by the full augmentation set: using just the top 1\% augmented data samples, as scored by the discriminator, is sufficient to generalize to previously unseen environments. Third, we train the discriminator using \textit{alignment-based} similarity metric that enables the model to align same concepts in the language and visual modalities. We provide a  qualitative assessment of the alignment learned by the model. Finally, we show that a navigation agent, when initialized with components of fully-trained discriminator, outperforms the existing benchmark on success rate by over 10\% relative measure on previously unseen environments.

\section{The Room-to-Room Dataset}
\label{sec:data}

Room-to-Room (R2R) is a visually-grounded natural language navigation dataset in photo-realistic environments \cite{Anderson:2018:VLN}. Each environment is defined by a graph where nodes are locations with egocentric panoramic images and edges define valid connections for agent navigation. The navigation dataset consists of language instructions paired with reference paths, where each path is defined by a sequence of graph nodes. The data collection process is based on sampling pairs of start/end nodes and defining the shortest path between the two. Furthermore the collection process ensures no paths are shorter than 5m and must be between 4 to 6 edges. Each sampled path is associated with 3 natural language instructions collected from Amazon Mechanical Turk with an average length of 29 tokens from a vocabulary of 3.1k tokens. Apart from the training set, the dataset includes two validation sets and a test set. One of the validation sets includes new instructions on environments overlapping with the training set (Validation Seen), and the other is entirely disjoint from the training set (Validation Unseen).

Several metrics are commonly used to evaluate agents' ability to follow navigation instructions. \textit{Path Length (PL)} measures the total length of the predicted path, where the optimal value is the length of the reference path. \textit{Navigation Error (NE)} measures the distance between the last nodes in the predicted path and the reference path. \textit{Success Rate (SR)} measures how often the last node in the predicted path is within some threshold distance $d_{th}$ of the last node in the reference path. More recently, \citet{Anderson:2018:Evaluation} proposed the \textit{Success weighted by Path Length (SPL)} measure that also considers whether the success criteria was met (i.e., whether the last node in the predicted path is within some threshold $d_{th}$ of the reference path) and the normalized path length. Agents should minimize NE and maximize SR and SPL.

\section{Discriminator Model}
\label{sec:discriminator}

VLN tasks are composed of instruction-path pairs, where a path is a sequence of connected locations along with their corresponding perceptual contexts in some environment. While the core task is to create agents that can follow the navigation instructions to reproduce estimates of reference paths, we instead explore models that focus on the simpler problem of judging whether an instruction-path pair are a good match for one another. These models would be useful in measuring the quality of machine-generated instruction-path pairs. Another reasonable expectation from such models would be that they are also able to align similar concepts in the two modalities, \eg, in an instruction like ``\textit{Turn right and move forward around the bed, enter the bathroom and wait there.}'', it is expected that the word \textit{bed} is better aligned with a location on the path that has a bed in the agent's egocentric view.

To this effect, we train a discriminator model that learns to delineate positive instruction-path pairs from negative pairs sampled using different strategies described in Sec.\ref{subsec:discriminator_training}. The discrimination is based on an alignment-based similarity score that determines how well the two input sequences align. This encourages the model to map perceptual and textual signals for final discrimination.

\subsection{Model Structure}
\label{subsec:disc_structure}

We use a two-tower architecture to independently encode the two sequences, with one tower encoding the token sequence $x_1, x_2, ..., x_n$ in the instruction $\mathcal{X}$ and another tower encoding the visual input sequence $v_1, v_2, ..., v_m$ from the path $\mathcal{V}$. Each tower is a bi-directional LSTM \cite{Schuster1997BidirectionalRN} which constructs the latent space representation $H$ of a sequence $i_1, i_2, ..., i_k$ following:

\begin{small}
\begin{align}
    H &= [h_1; h_2; ...; h_k]  \\
    h_t &= g(\overrightarrow{h}_t, \overleftarrow{h}_t)  \\
    \overrightarrow{h}_t &= LSTM(i_t, \overrightarrow{h}_{t-1})  \\
    \overleftarrow{h}_t &= LSTM(i_t, \overleftarrow{h}_{t+1})
\end{align}
\end{small}

\noindent
where $g$ function is used to combine the output of forward and backward LSTM layers. In our implementation, $g$ is the concatenation operator.

We denote the latent space representation of instruction $\mathcal{X}$ as $H^X$ and path $\mathcal{V}$ as $H^V$ and compute the alignment-based similarity score as following:

\begin{small}
\begin{align}
    A &= H^X({H^V})^T  \label{eq:A_compute} \\
    \{c\}_{l=1}^{l=X} &= \text{softmax}(A^l) \cdot A^l  \label{eq:A_softmax} \\
    \text{score} &= \text{softmin}(\{c\}_{l=1}^{l=X}) \cdot \{c\}_{l=1}^{l=X} \label{eq:A_softmin} 
\end{align}
\end{small}

\noindent
where $(.)^T$ is matrix transpose transformation, $A$ is the alignment matrix whose dimensions are $[n, m]$, $A^l$ is the $l$-th row vector in $A$ and $\text{softmin(Z)} = \frac{\exp^{-Z_{j}}}{\sum{\exp^{-Z_{j}}}}$. Eq.\ref{eq:A_softmax} corresponds to taking a softmax along the columns and summing the columns, which amounts to content-based pooling across columns. Then we apply softmin operation along the rows and sum the rows up to get a scalar in Eq.\ref{eq:A_softmin}. Intuitively, optimizing this score encourages the learning algorithm to construct the best worst-case sequence alignment between the two input sequences in latent space.

\subsection{Training}
\label{subsec:discriminator_training}

Training data consists of instruction-path pairs which may be similar (positives) or dissimilar (negatives). The training objective maximizes the log-likelihood of predicting higher alignment-based similarity scores for similar pairs.

We use the human annotated demonstrations in the R2R dataset as our positives and explore three strategies for sampling negatives. For a given instruction-path pair, we sample negatives by keeping the same instruction but altering the path sequence by:

\begin{itemize}
    \item \textit{Path Substitution (PS)} -- randomly picking other paths from the same environment as negatives.
    \item \textit{Partial Reordering (PR)} -- keeping the first and last nodes in the path unaltered and shuffling the intermediate locations of the path.
    \item \textit{Random Walks (RW)} -- sampling random paths of the same length as the original path that either (1) start at the same location and end sufficiently far from the original path or (2) end at the same location and start sufficiently far from the original path.
\end{itemize}

\begin{table}
\centering
\begin{tabular}{llll|r}
Learning          & PS        & PR        & RW         & AUC \\
\hline
no-curriculum     & \cmark    &           &            & 64.5 \\
no-curriculum     &           & \cmark    &            & 60.5 \\
no-curriculum     &           &           & \cmark     & 63.1 \\
no-curriculum     & \cmark    & \cmark    &            & 72.1 \\
no-curriculum     &           & \cmark    & \cmark     & 66.0 \\ 
no-curriculum     & \cmark    &           & \cmark     & 70.8 \\
no-curriculum     & \cmark    & \cmark    & \cmark     & 72.0 \\
\hline
curriculum        & \cmark    & \cmark    & \cmark     & \textbf{76.2} \\
\end{tabular}
\caption{Results on training in different combinations of datasets and evaluating against validation dataset containing PR and RW negatives only.\label{tab:sampling-strategy}}
\end{table}

\section{Results}
\label{sec:results}

Our experiments are conducted using the R2R dataset \cite{Anderson:2018:VLN}. Recently, \citet{Fried:2018:Speaker} introduced an augmented dataset (referred to as {\friedaug} from now on) that is generated by using a speaker model and they show that the models trained with both the original data and the machine-generated augmented data improves agent success rates. 

We show three main results. First, the discriminator effectively differentiates between high-quality and low-quality paths in {\friedaug}. Second, we rank all instruction-path pairs in {\friedaug} with the discriminator and train with a small fraction judged to be the highest quality---using just the top 1\% to 5\% (the highest quality pairs) provides most of the benefits derived from the entirety of {\friedaug} when generalizing to previously unseen environments. Finally, we initialize a navigation agent with the visual and language components of the trained discriminator. This strategy allows the agent to benefit from the discriminator's multi-modal alignment capability and more effectively learn from the human-annotated instructions. This agent outperforms existing benchmarks on previously unseen environments as a result.


\begin{figure}
\includegraphics[width=\linewidth]{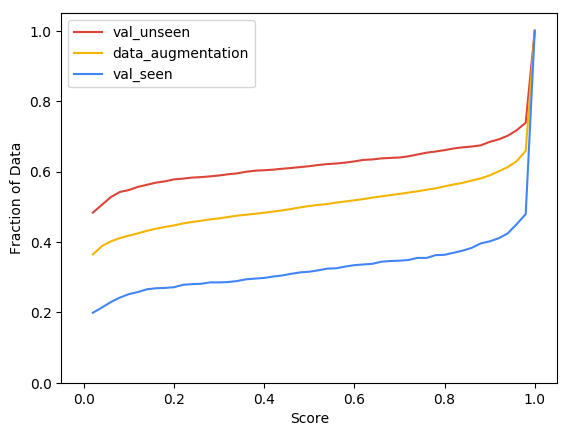}
\caption{Culmulative distributions of discriminator scores for different datasets. The mean of distribution for R2R \textit{validation seen}, \textit{{\friedaug}} and R2R \textit{validation unseen} is 0.679, 0.501, and 0.382 respectively.}
\label{fig:discriminator-distribution}
\end{figure}

\subsection{Discriminator Results}

We create two kinds of dataset for each of the negative sampling strategies defined in Section \ref{subsec:discriminator_training} -- a training set from paths in R2R \textit{train} split and validation set from paths in R2R \textit{validation seen} and \textit{validation unseen} splits. The area-under ROC curve (AUC) is used as the evaluation metric for the discriminator. From preliminary studies, we observed that the discriminator trained on dataset containing PS negatives achieved AUC of 83\% on validation a dataset containing PS negatives only, but fails to generalize to validation set containing PR and PW negatives (AUC of 64.5\%). This is because it is easy to score PS negatives by just attending to first or last locations, while scoring PR and PW negatives may require carefully aligning the full sequence pair. Therefore, to keep the task challenging, the validation set was limited to contain validation splits from PR and RW negative sampling strategies only. Table \ref{tab:sampling-strategy} shows the results of training the discriminator using various combinations of negative sampling.

\begin{table*}[!ht]
\centering
\begin{tabular}{c|c|c}

Dataset & Score & Example \\
\hline
\multirow{2}{*}{\shortstack[1]{Fried- \\ Augmented}} & 0.001  & \includegraphics[width=0.7\linewidth, trim={0 3cm 0 0},clip]{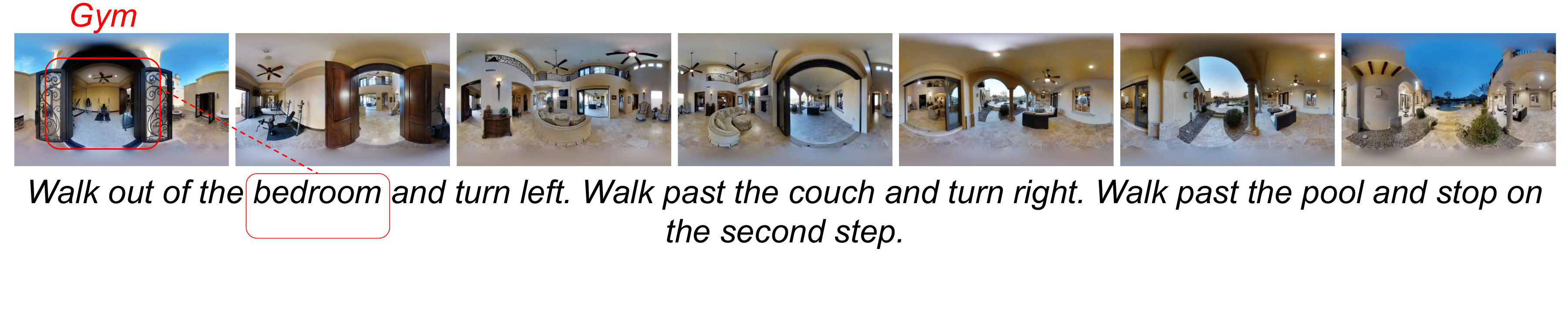} \\
& 0.999 & \includegraphics[width=0.7\linewidth, trim={0 4cm 0 0},clip]{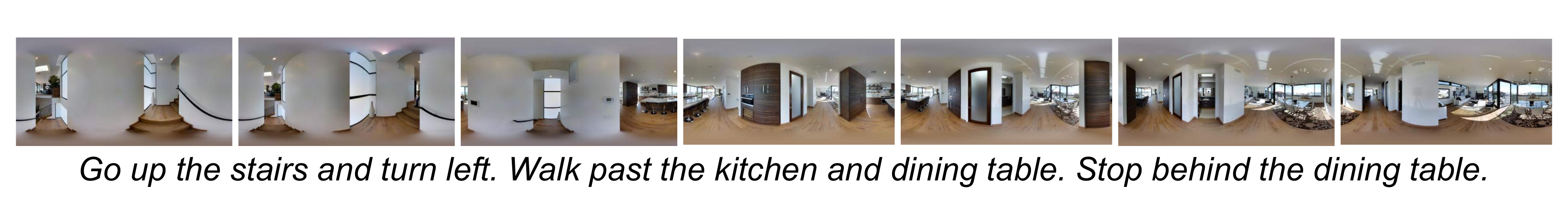} \\
\hline
\multirow{2}{*}{\shortstack[1]{Validation \\ Seen}} & 0.014 & \includegraphics[width=0.7\linewidth, trim={0 4cm 0 0},clip]{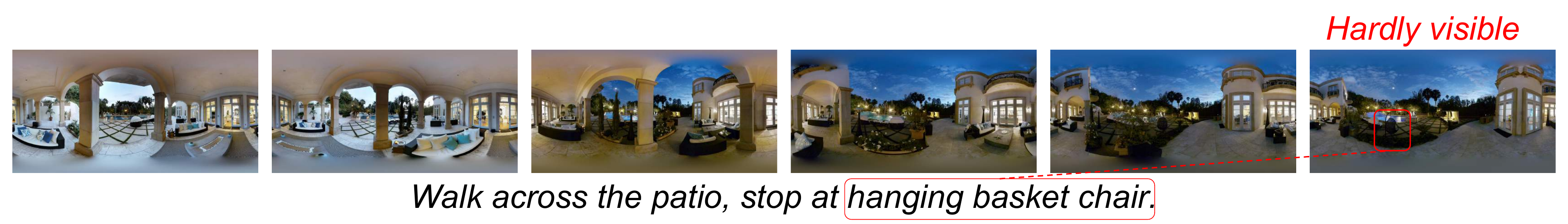} \\
& 0.999 & \includegraphics[width=0.7\linewidth, trim={0 4cm 0 0},clip]{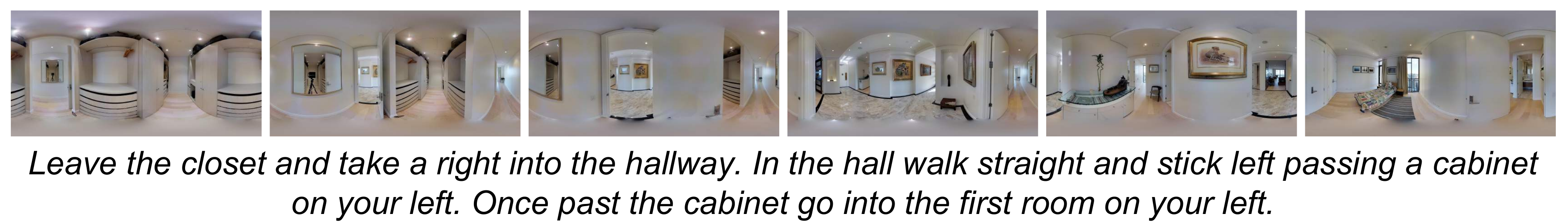} \\
\hline
\multirow{2}{*}{\shortstack[1]{Validation \\ Unseen}} & 0.00004 & \includegraphics[width=0.7\linewidth, trim={0 3cm 0 0},clip]{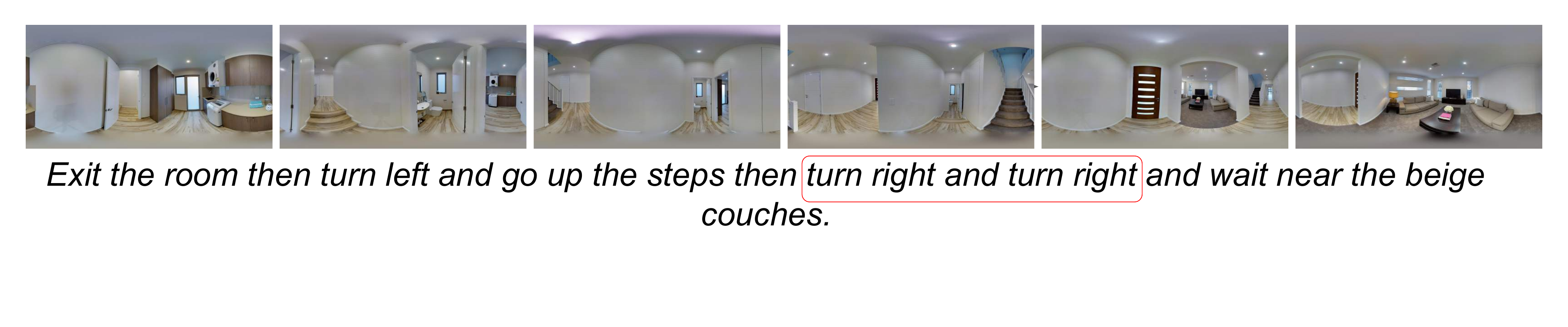} \\
& 0.9808 & \includegraphics[width=0.7\linewidth, trim={0 4cm 0 0},clip]{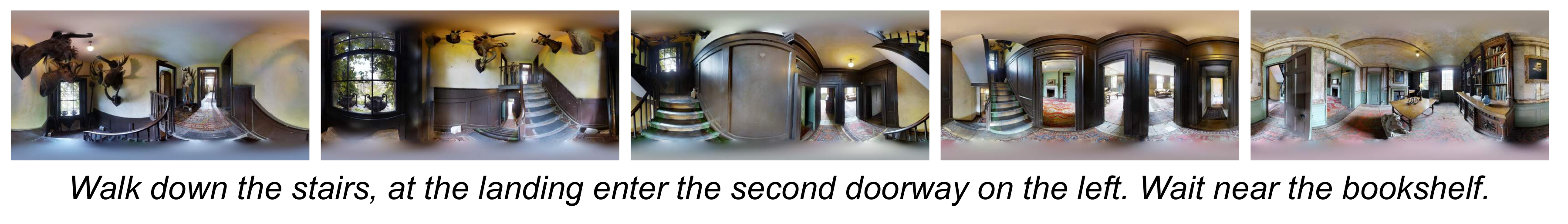} \\
\end{tabular}

\caption{Selected samples from datasets with discriminator scores.} \label{tab:score-viz}

\end{table*}

Generally, training the discriminator with PS negatives helps model performance across the board. Simple mismatch patterns in PS negatives help bootstrap the model with a good initial policy for further fine-tuning on tougher negatives patterns in PR and RW variations. For example in PS negatives, a path that starts in a bathroom does not match with an instruction that begins with ``\textit{Exit the bedroom.}''--this would be an easy discrimination pair. In contrast, learning from just PR and RW negatives fails to reach similar performance. To further confirm this hypothesis, we train a discriminator using curriculum learning \cite{Bengio:2009:CL} where the model is first trained on only PS negatives and then fine-tuned on PR and RW negatives. This strategy outperforms all others, and the resulting best performing discriminator is used for conducting studies in the following subsections.


\textbf{Discriminator Score Distribution}
Fig.\ref{fig:discriminator-distribution} shows the discriminator's score distribution on different R2R datasets. Since {\friedaug} contains paths from houses seen during training, it would be expected that discriminator's scores on \textit{validation seen} and {\friedaug} datasets be the same if the data quality is comparable. However there is a clear gap in the discriminator's confidence between the two datasets. This matches our subjective analysis of {\friedaug} where we observed many paths had clear starting/ending descriptions but the middle sections were often garbled and had little connection to the perceptual path being described. Table \ref{tab:score-viz} contains some samples with corresponding discriminator scores.

\begin{table*}
\centering
\begin{tabular}{l|l|rr|rr|rr|rr}
Dataset size & Strategy  & \multicolumn{2}{l}{PL} & \multicolumn{2}{l}{NE $\downarrow$} & \multicolumn{2}{l}{SR $\uparrow$} & \multicolumn{2}{l}{SPL $\uparrow$} \\
\hline
 & & U & S & U & S & U & S & U & S\\
\hline 
\multirow{5}{*}{1\%}      & Top                &  11.2 & 11.1 & 8.5 & 8.5 & 20.4 & 21.2 & 16.6 & \textbf{17.6} \\
                       & Bottom                &  10.8 & 10.7 & 8.9 & 9.0 & 15.4 & 16.3 & 14.1 & 13.1 \\
                       & Random Full           &  11.7 & 12.5 & 8.1  & 8.3 & 22.1 & 21.2 & \textbf{17.9} & 16.6 \\
                       & Random Bottom       &  14.2 & 15.8 & 8.4 & 8.1 & 19.7 & 21.7 & 14.3 & 15.6 \\
                       & Random Top          &  15.9 & 15.6 & \textbf{7.9} & \textbf{7.6} & \textbf{22.6} & \textbf{25.4} & 15.2 & 14.8 \\
\hline 
\multirow{5}{*}{2\%}   & Top                   &  11.3 & 11.7 & 8.2 & 7.9 & 22.3 & 25.5 & 18.5 & 21.0 \\
                       & Bottom                &  11.4 & 14.5 & 8.4 & 9.1 & 17.5 & 17.7 & 14.1 & 12.7 \\
                       & Random Full           &  13.3 & 10.8 & 7.9 & 7.9 & 24.3 & 25.5 & 18.2 & \textbf{22.7} \\
                       & Random Bottom       &  15.2 & 18.2 & 8.1 & 8.1 & 20.5 & 20.8 & 11.8 & 16.0  \\
                       & Random Top          &  12.9 & 14.0 & \textbf{7.6} & \textbf{7.5} & \textbf{25.6} & \textbf{25.8} & \textbf{19.5} & 19.7 \\
\hline
\multirow{5}{*}{5\%}   & Top                   & 17.6 & 16.9 & 7.7  & 7.2 & 24.6 & 28.2 & 14.4 & 18.2  \\
                       & Bottom                & 10.0 & 10.2 & 8.3  & 8.2 & 20.1 & 23.2 & \textbf{17.1} & 19.4 \\
                       & Random Full               & 17.8 & 21.4 & 7.3  & 7.0  & 27.2 & 29.1 & 16.4 & 14.3 \\
                       & Random Bottom       & 16.3 & 10.4 & 7.9  & 8.3 & 22.1 & 23.0 & 14.2 & 20.1  \\
                       & Random Top          & 20.0 & 15.0 & \textbf{7.0} & \textbf{6.9} & \textbf{27.7} & \textbf{30.6} & 14.8 & \textbf{22.1} \\
\end{tabular}
\caption{Results on R2R validation unseen paths (U) and seen paths (S) when trained only with small fraction of {\friedaug} ordered by discriminator scores. For \textit{Random Full} study, examples are sampled uniformly over entire dataset. For \textit{Random Top/Bottom} study, examples are sampled from top/bottom 40\% of ordered dataset. SPL and SR are reported as percentages and NE and PL in meters. \label{tab:r2r-small-dataset}}
\end{table*}

Finally we note that the discriminator scores on \textit{validation unseen} are rather conservative even though the model differentiates between positives and negatives from validation set reasonably well (last row in Table \ref{tab:sampling-strategy}).


\subsection{Training Navigation Agent}
\label{subsec:train_nav_agent}
We conducted studies on various approaches to incorporate selected samples from {\friedaug} to train navigation agents and measure their impact on agent navigation performance. The studies illustrate that navigation agents have higher success rates when they are trained on higher-quality data (identified by discriminator) with sufficient diversity (introduced by random sampling). When the agents are trained with mixing selected samples from {\friedaug} to R2R \textit{train} dataset, only the top 1\% from {\friedaug} is needed to match the performance on existing benchmarks.

\begin{table*}
\centering
\begin{tabular}{l|rr|rr|rr|rr}

Dataset  &  \multicolumn{2}{l}{PL} & \multicolumn{2}{l}{NE $\downarrow$} & \multicolumn{2}{l}{SR $\uparrow$} & \multicolumn{2}{l}{SPL $\uparrow$} \\
\hline
 & U & S & U & S & U & S & U & S\\
\hline
Benchmark\footnote[2] & - & - &  6.6 & 3.36 & 35.5 & 66.4 & - & - \\
\hline
0\%  &  17.8 & 18.5 & 6.8 & 5.3  & 32.1 & 46.1 & 21.9 & 30.3\\
1\%  &  12.5 & 11.2 & 6.4 & 5.7  & 35.2 & 45.3 & 28.9 & 39.1 \\
2\%        &  14.5 & 15.1 & 6.5 & 5.5  & 35.7 & 44.6 & 27.0 & 34.1 \\
5\%        &  17.0 & 12.9 & \textbf{6.1} & 5.6 & 36.0 & 44.8 & 23.6 & 37.0 \\
40\%        &  14.9 & 11.9 & 6.4 & 5.5 & \textbf{36.5} & 49.1 & 27.1 & 43.4 \\
60\% & 16.8 & 15.7 & 6.3 & 5.3 & 36.0 & 47.2 & 24.7 & 35.4 \\
80\% & 17.1 & 18.5 & 6.2 & 5.2 & 35.8 & 45.0 & 23.8 & 29.6 \\
100\% &  15.6 & 15.9 & 6.4 & \textbf{4.9} & 36.0 & \textbf{51.9} & \textbf{29.0} & \textbf{43.0} \\
\end{tabular}
\caption{Results\footnote[3] - on R2R validation unseen (U) and validation seen (S) paths when trained with full training set and selected fraction of {\friedaug}. SPL and SR are reported as percentages and NE and PL in meters. \label{tab:r2r-golden-augmentation-dataset}}
\end{table*}

\paragraph{Training Setup.} The training setup of the navigation agent is identical to \citet{Fried:2018:Speaker}. The agent learns to map the natural language instruction $\mathcal{X}$ and the initial visual scene $v_1$ to a sequence of actions $a_{1..T}$. Language instructions $\mathcal{X}=x_{1..n}$ are initialized with pre-trained GloVe word embeddings \cite{Pennington:2014:GloVe} and encoded using a bidirectional RNN \cite{Schuster1997BidirectionalRN}. At each time step $t$, the agent perceives a 360-degree panoramic view of its surroundings from the current location. The view is discretized into $m$ view angles ($m=36$ in our implementation, 3 elevations x 12 headings at 30-degree intervals). The image at view angle $i$, heading angle $\phi$ and elevation angle $\theta$ is represented by a concatenation of the pre-trained CNN image features with the 4-dimensional orientation feature [sin $\phi$; cos $\phi$; sin $\theta$; cos $\theta$] to form $v_{t,i}$. As in \citet{Fried:2018:Speaker}, the agent is trained using \textit{student forcing} where actions are sampled from the model during training, and supervised using a shortest-path action to reach the goal state.

\paragraph{Training using {\friedaug} only.} The experiments in Table \ref{tab:r2r-small-dataset} are based on training a navigation agent on different fractions of the {\friedaug} dataset (X=$\{1\%, 2\%, 5\%\}$) and sampling from different parts of the discriminator score distribution (\textit{Top}, \textit{Bottom}, \textit{Random Full}, \textit{Random Top}, \textit{Random Bottom}). The trained agents are evaluated on both \textit{validation seen} and \textit{validation unseen} datasets.

Not surprisingly, the agents trained on examples sampled from the \textit{Top} score distribution consistently outperform the agents trained on examples from the \textit{Bottom} score distribution. Interestingly, the agents trained using the \textit{Random Full} samples is slightly better than agents trained using just the \textit{Top} samples. This suggests that the agent benefits from higher diversity samples. This is confirmed by the study \textit{Random Top} where the agents trained using high quality samples with sufficient diversity consistently outperform all other agents.

\begin{table*}
\centering
\begin{tabular}{l|l|l|l|l|l}

Method & Split &  PL & NE $\downarrow$ & SR $\uparrow$ & SPL $\uparrow$ \\
\hline
\multirow{2}{*}{Speaker-Follower model \cite{Fried:2018:Speaker}} & U & - & 6.6 & 35.5 & - \\
 & S & - & \textbf{3.36} & \textbf{66.4} & - \\
\hline
\multirow{2}{*}{\shortstack[1]{Speaker-Follower model (our implementation)}} & U & 15.6 & {6.4} & {36.0} & 29.0 \\
 & S & 15.9 & 4.9 & 51.9 & 43.0 \\
\hline
\multirow{2}{*}{\shortstack[1]{Our implementation, using discriminator pre-training}} & U & 16.7 & \textbf{5.9} & \textbf{39.1} & 26.8 \\
 & S & 15.4 & 5.0 & 50.4 & 39.1 \\

\end{tabular}
\caption{Results on R2R validation unseen (U) and validation seen (S) paths after initializing navigation agent's instruction and visual encoders with discriminator.\label{tab:pretrained-encoder}}
\end{table*}

\paragraph{Training using both R2R \textit{train} and {\friedaug}.} To further investigate the utility of the discriminator, the navigation agent is trained with the full R2R \textit{train} dataset (which contains human annotated data) as well as selected fractions of {\friedaug}\footnote{We tried training on {\friedaug} first and then fine-tuning on R2R \textit{train} dataset, as done in \citet{Fried:2018:Speaker}, but didn't find any appreciable difference in agent's performance in any of the experiments.}. Table \ref{tab:r2r-golden-augmentation-dataset} shows the results.

\textit{Validation Unseen}: The performance of the agents trained with just 1\% {\friedaug} matches with benchmark for NE and SR. With just 5\% {\friedaug}, the agent starts outperforming the benchmark for NE and SR. Since {\friedaug} was generated by a speaker model that was trained on R2R \textit{train}, the language diversity in the dataset is limited, as evidenced by the unique token count: R2R \textit{train} has 2,602 unique tokens while {\friedaug} has only unique 369 tokens. The studies show that only a small fraction of top scored {\friedaug} is needed to augment R2R \textit{train} to achieve the full performance gain over the benchmark.

\begin{figure*}[!ht]
\includegraphics[width=\textwidth]{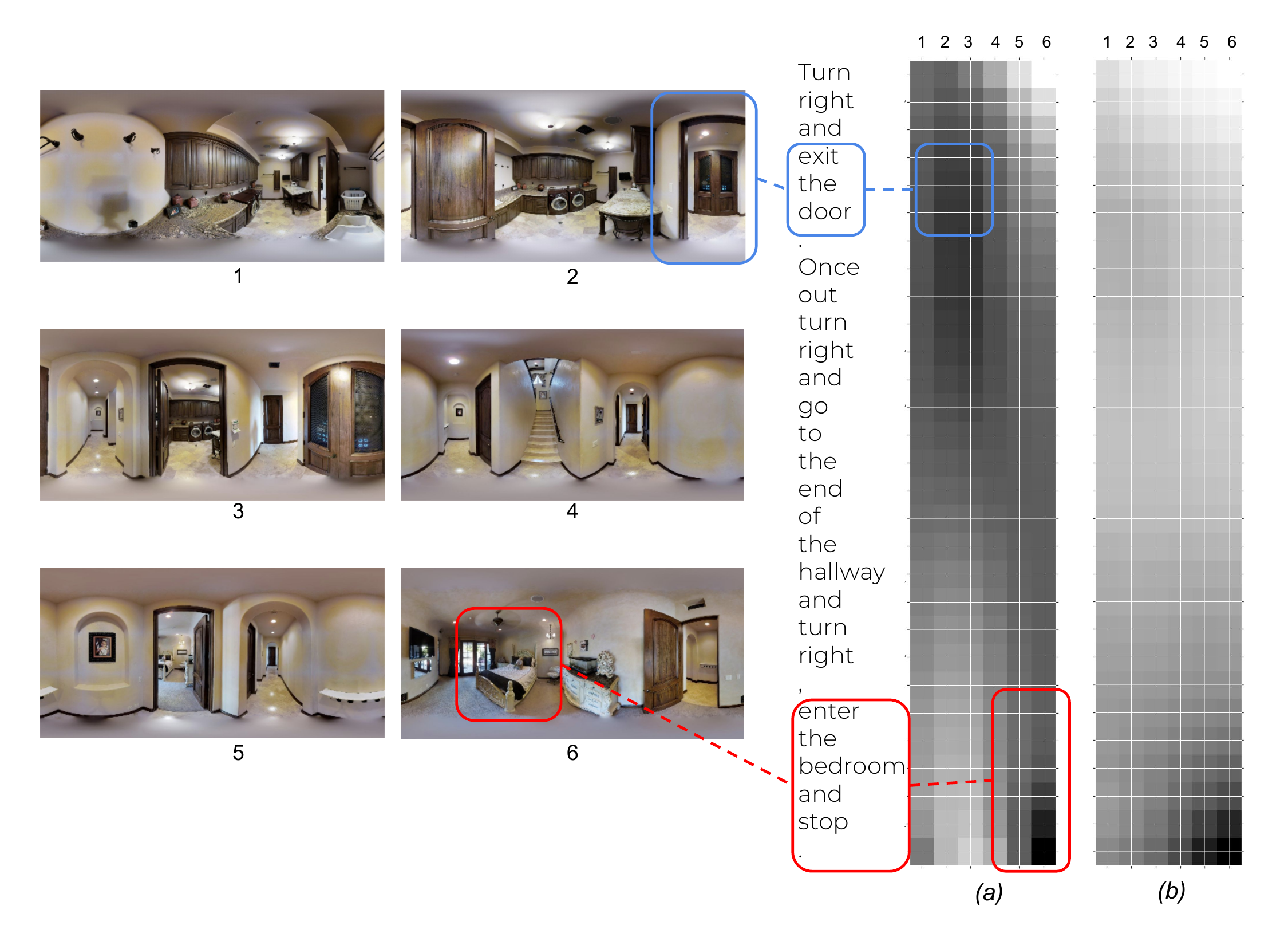}
\caption{Alignment matrix (Eq.\ref{eq:A_compute}) for discriminator model trained (a) with curriculum learning on the dataset containing PS, PR, RW negatives (b) without curriculum learning on the dataset with PS negatives only. Note that darker means higher alignment. \label{fig:curriculum-learning-attention}}
\end{figure*}

\textit{Validation Seen}: Since {\friedaug} contains paths from houses seen during training, mixing more of it with R2R \textit{train} helps the agent overfit on \textit{validation seen}. Indeed, the model's performance increases nearly monotonically on NE and SR as higher fraction of {\friedaug} is mixed in the training data. The agent performs best when it is trained on all of {\friedaug}.

\footnotetext[2]{For a fair comparison, the benchmark is the Speaker-Follower model from \citet{Fried:2018:Speaker} which uses panoramic action space and augmented data, but no beam search (pragmatic inference).}
\footnotetext[3]{Our results of the agents trained on the full R2R \textit{train} and 100\% {\friedaug} match with Speaker-Follower benchmark on validation unseen but are lower on validation seen. This is likely due to differences in model capacity, hyper-parameter choices and image features used in our implementation. The image features used in our implementation are obtained through a convolutional network trained with a semantic ranking objective on a proprietary image dataset with over 100+ million images \cite{Wang:2014:Learning}.}

\textbf{Initializing with Discriminator}. To further demonstrate the usefulness of the discriminator strategy, we initialize a navigation agent's instruction and visual encoder using the discriminator's instruction and visual encoder respectively. We note here that since the navigation agent encodes the visual input sequence using LSTM, we re-train the best performing discriminator model using LSTM (instead of bidirectional-LSTM) visual encoder so that the learned representations can be transferred correctly without any loss of information. We observed a minor degradation in the performance of the modified discriminator. The navigation agent so initialized is then trained as usual using \textit{student forcing}. The agent benefits from the multi-modal alignment learned by the discriminator and outperforms the benchmark on the \textit{Validation Unseen} set, as shown in Table \ref{tab:pretrained-encoder}. This is the condition that best informs how well the agent generalizes. Nevertheless, performance drops on \textit{Validation Seen}, so further experimentation will hopefully lead to improvements on both.

\subsection{Visualizing Discriminator Alignment}
\label{subsec:visual_attention}
We plot the alignment matrix $A$ (Eq.\ref{eq:A_compute}) from the discriminator for a given instruction-path pair to try to better understand how well the model learns to align the two modalities as hypothesized.
As a comparison point, we also plot the alignment matrix for a model trained on the dataset with PS negatives only.
As discussed before, it is expected that the discriminator trained on the dataset containing only PS negatives tends to exploit easy-to-find patterns in negatives and make predictions without carefully attending to full instruction-path sequence.

Fig.\ref{fig:curriculum-learning-attention} shows the difference between multi-modal alignment for the two models. While there is no clear alignment between the two sequences for the model trained with PS negatives only (except maybe towards the end of sequences, as expected), there is a visible diagonal pattern in the alignment for the best discriminator. In fact, there is appreciable alignment at the correct positions in the two sequences, \eg, the phrase \textit{exit the door} aligns with the image(s) in the path containing the object \textit{door}, and similarly for the phrase \textit{enter the bedroom}.


\section{Related Work}
\label{sec:related}

The release of Room-to-Room (R2R for short) dataset \cite{Anderson:2018:VLN} has sparked research interest in multi-modal understanding. The dataset presents a unique challenge as it not only substitutes virtual environments (\eg, \citet{Macmahon06walkthe}) with photo-realistic environments but also describes the paths in the environment using human-annotated instructions (as opposed to formulaic instructions provided by mapping applications \eg, \citet{Cirik:2018:StreetView}). A number of methods \cite{Anderson:2018:VLN,Fried:2018:Speaker,Wang:2018:RCM,Ma:2019:SelfMonitoringAgent,Wang2018Look,ma2019regretful} have been proposed recently to solve the navigation task described in R2R dataset. All these methods build models for agents that learn to navigate in R2R environment and are trained on the entire R2R dataset as well as the augmented dataset introduced by \citet{Fried:2018:Speaker} which is generated by a speaker model trained on human-annotated instructions. 

Our work is inspired by the idea of Generative Adversarial Nets \cite{NIPS2014_5423}, which use a discriminative model to discriminate real and fake distribution from generative model. We propose models that learn to discriminate between high-quality instruction-path pairs from lower quality pairs. This discriminative task becomes important for VLN challenges as the data is usually limited in such domains and data augmentation is a common trick used to overcome the shortage of available human-annotated instruction-path pairs. While all experiments in this work focus on R2R dataset, same ideas can easily be extended to improve navigation agents for other datasets like Touchdown \cite{ChenTouchdown2018}.

\section{Conclusion}
\label{sec:conclusion}

We show that the discriminator model is capable of differentiating high-quality examples from low-quality ones in machine-generated augmentation to VLN datasets. The discriminator when trained with \textit{alignment based} similarity score on cheaply mined negative paths learns to align similar concepts in the two modalities. The navigation agent when initialized with the discriminator generalizes to instruction-path pairs from previously unseen environments and outperforms the benchmark.

For future work, the discriminator can be used in conjunction with generative models producing extensions to human-labeled data, where it can filter out low-quality augmented data during generation as well as act as a reward signal to incentivize generative model to generate higher quality data. The multi-modal alignment learned by the discriminator can be used to segment the instruction-path pair into several shorter instruction-path pairs which can then be used for creating a curriculum of easy to hard tasks for the navigation agent to learn on. It is worth noting that the trained discriminator model is general enough to be useful for any downstream task which can benefit from such multi-modal alignment measure and not limited to VLN task that we use in this work.

\textbf{}

\end{document}